\documentclass[11pt]{article}

\usepackage[final]{acl}

\usepackage{times}
\usepackage{latexsym}

\usepackage[T1]{fontenc}

\usepackage[utf8]{inputenc}

\usepackage{microtype}

\usepackage{inconsolata}

\usepackage{graphicx}
\usepackage{amsmath}
\usepackage{amssymb}
\usepackage{booktabs}
\usepackage{multirow}
\usepackage{colortbl}
\usepackage{pifont}

\title{GraMRAG: Orchestrating Multi-Agent Multi-Step Reasoning via Graph Memory with Reinforcement Learning}

\author{Zhongyu Wang\textsuperscript{\textdagger} \\
  Beihang University \\
  \texttt{wangzhongyu@buaa.edu.cn} \\}

\begin{document}
\maketitle

\renewcommand{\thefootnote}{\fnsymbol{footnote}}
\footnotetext[2]{Corresponding author.}

\begin{abstract}
Although existing multi-agent Retrieval-Augmented Generation (RAG) systems have demonstrated promise on complex multimodal reasoning tasks, they remain fundamentally limited in reasoning depth and memory structure, suffering from inadequate retrieval and state blindness when answering knowledge-intensive questions. To address these limitations, we propose GraMRAG, a graph memory-guided multi-agent RAG framework that integrates a dynamic multimodal memory graph to enable stable, multi-step multimodal reasoning. We introduce a vision-text bridged reasoning paradigm that unifies multi-scale entity cropping with a ReAct-style visual toolchain, enhancing the long-horizon cross-modal reasoning capability. We further construct a multimodal memory graph that formalizes agent reasoning as a dynamic directed acyclic graph (DAG), explicitly modeling action-observation dependencies to mitigate state blindness and suppress redundant retrieval. Moreover, we propose Topology-Aware Policy Optimization (TAPO) that leverages graph topology for critical path identification and targeted node pruning, enabling fine-grained credit assignment across multi-step reasoning trajectories. Extensive experiments on challenging multimodal benchmarks demonstrate that our approach consistently outperforms existing baselines and achieves state-of-the-art performance on complex long-horizon reasoning tasks.
\end{abstract}

\section{Introduction}

\begin{figure}[t]
  \centering
  \includegraphics[width=0.95\columnwidth]{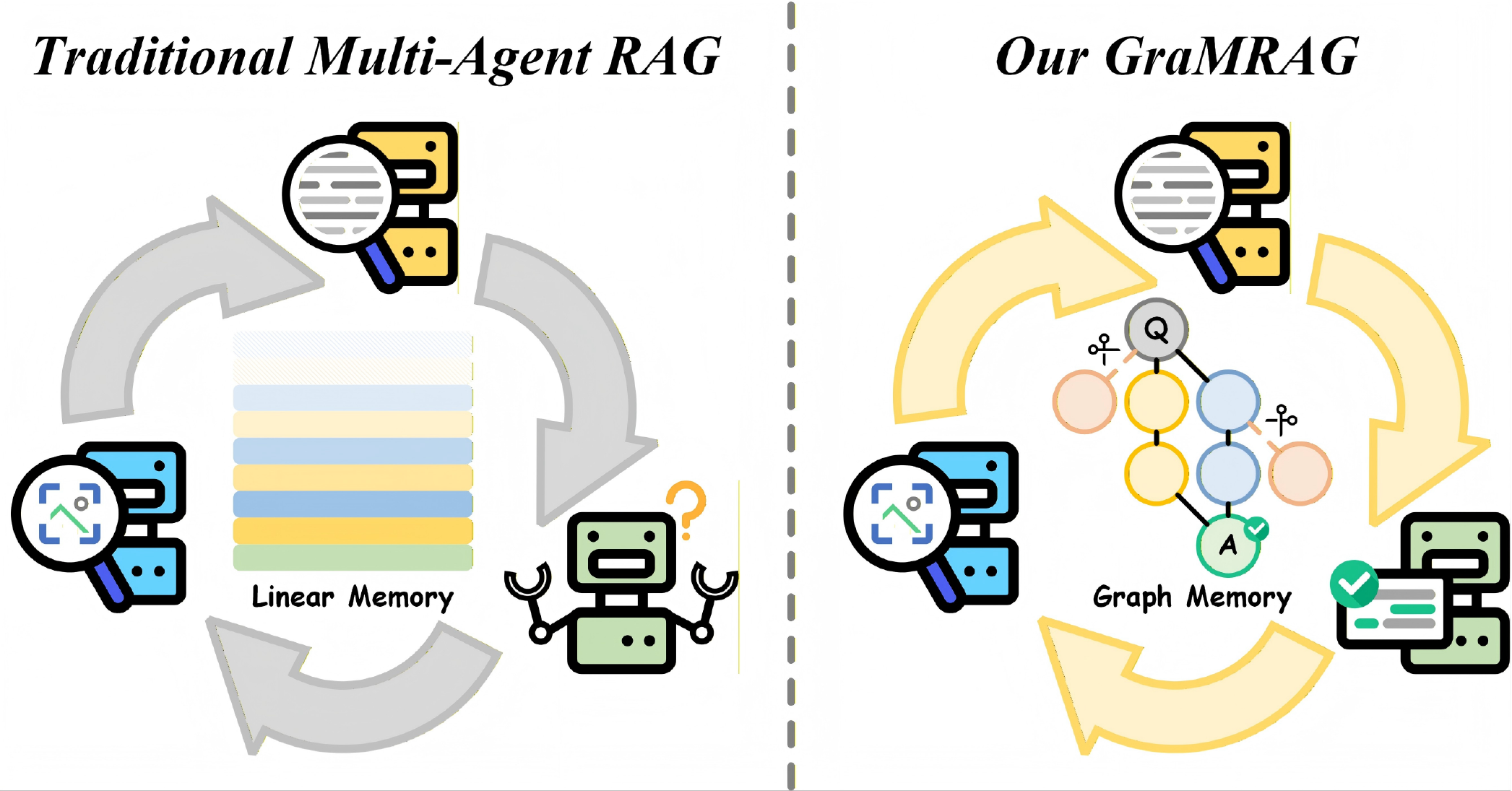}
  \caption{Comparison between our GraMRAG and existing multi-agent RAG systems.}
  \label{fig:teaser}
\end{figure}

{\noindent\itshape
“Never memorize something that you can look up.”%
\begin{flushright}
\normalfont--- Albert Einstein
\end{flushright}
}

Retrieval-Augmented Generation (RAG) extends the knowledge boundaries of Multimodal Large Language Models (MLLMs) by retrieving relevant information from external knowledge bases~\cite{wang2026mdocrag, fan2026segmem, hu2026cog}. Recent advances in MLLMs have convincingly demonstrated that multimodal agentic RAG systems are capable of efficient retrieval and reasoning over large-scale corpora comprising interleaved textual and visual content~\cite{zhang2026patho, guan2025kg}. MLLM-based multi-agent architectures, leveraging specialized collaboration among components, exhibit superior performance over single-agent frameworks on complex multimodal RAG tasks~\cite{zhou2026mobile, chang2025main, xu2025multiagentesc}.

However, existing multi-agent RAG systems face three key limitations. 
First, the average number of retrieval rounds in current systems remains limited, predisposing agents to prematurely terminate exploration and settle for partially available evidence on complex tasks~\cite{jiang2025mmsearch, wu2025mmsearch, team2025tongyi}. This fundamentally impedes the transfer of the long-horizon planning capability from text-based deep research to multimodal scenarios, rendering these systems inadequate for tens of iterative reasoning steps demanded by real-world deep research. Second, prevailing linear or iterative summarization-based memory structures~\cite{yao2022react, zhou2025mem1, xu2025mem, wu2025resum}, lacking explicit modeling of agent reasoning states, are prone to context expands and state blindness as interaction rounds accumulate, resulting in repetitive queries and redundant retrieval in multi-hop tasks. Third, conventional outcome-based sparse rewards coarsely propagate terminal-state signals across entire trajectories~\cite{huang2026vision, zhou2025mem1}, causing effective retrieval steps to be erroneously penalized in negative samples while redundant operations are undeservedly rewarded in positive ones, preventing fine-grained credit assignment across multi-step reasoning.

To address the mentioned limitations, we propose GraMRAG, a novel graph memory-guided multi-agent RAG framework. First, to bridge the long-horizon reasoning gap, we introduce a vision-text bridged reasoning paradigm that orchestrates multi-scale entity cropping with a ReAct-style visual toolchain, seamlessly transferring the iterative planning capability of text-only deep research models into the multimodal domain to enable long-horizon cross-modal reasoning trajectories. Second, to overcome state blindness and context expansion in linear memory paradigms, we construct a multimodal memory graph that formalizes the reasoning process as a dynamic directed acyclic graph (DAG), explicitly encoding agent actions and multimodal observations with their temporal and logical dependencies to mitigate redundant retrieval and state blindness. Third, to tackle the credit assignment problem from coarse terminal rewards, we introduce a Topology-Aware Policy Optimization (TAPO) algorithm that leverages the graph topology to identify critical reasoning paths and perform targeted node pruning, precisely masking false positive and false negative nodes during policy updates to enable fine-grained credit assignment across multi-step reasoning.

Our contributions are summarized as follows:

\begin{itemize}
\item We introduce a vision-text bridged reasoning paradigm that unifies multi-scale entity cropping with a ReAct-style visual toolchain, enabling long-horizon cross-modal reasoning.
\item We construct a multimodal memory graph that formalizes agent reasoning as a dynamic DAG to systematically mitigate state blindness and redundant retrieval.
\item We develop a TAPO algorithm leveraging graph topology for critical path identification and node pruning, enabling fine-grained credit assignment across multi-step reasoning.
\item We conduct extensive experiments on challenging multimodal benchmarks to evaluate our approach, demonstrating state-of-the-art performance over existing baselines.
\end{itemize}

\section{Related Work}
\subsection{Multi-Agent Systems in Multimodal RAG}
Multi-agent systems have demonstrated promise on complex multimodal reasoning tasks~\cite{wang2026mars, dong2026s, li2026draft, xinjie2025reagent, zhang2025belle}. By deploying multiple specialized agents that each focus on distinct facets of a problem, these architectures enable collaborative problem-solving that surpasses the capabilities of single-agent frameworks~\cite{ni2026multi, sinha2026concept, siingh2025getreason}. 
However, existing multi-agent RAG systems conduct insufficient retrieval rounds, causing agents to prematurely terminate exploration on complex tasks~\cite{geng2025webwatcher, narayan2025deepmmsearch}. Such shallow reasoning trajectories are fundamentally insufficient for demanding deep research scenarios that require tens of iterative reasoning steps. 
To address this challenge, we introduce a vision-text bridged reasoning paradigm integrating multi-scale entity cropping with a visual toolchain, using image descriptions as a cross-modal bridge to enable long-horizon planning in the multimodal domain.

\subsection{Memory Structures of Agent Systems}
The memory architectures of existing agent systems can broadly be categorized into two paradigms~\cite{zhang2025survey, ye2025agentfold}: the history accumulation paradigm, exemplified by ReAct~\cite{yao2022react}, which linearly concatenates all interaction records, and the iterative summarization paradigm, exemplified by Mem1~\cite{zhou2025mem1}, which controls context length by compressing observations. However, as the number of interaction rounds grows, both paradigms cause agents to progressively lose track of historical retrieval paths, a phenomenon we term state blindness, which in turn induces redundant queries and repetitive retrieval in multi-hop tasks. To address this issue, we construct a multimodal memory graph that formalizes the reasoning process as a dynamic DAG, explicitly encoding the temporal and logical dependencies between actions and observations, thereby effectively mitigating state blindness and redundant retrieval. 

\begin{figure*}[t]
    \centering
    \includegraphics[width=0.9\textwidth]{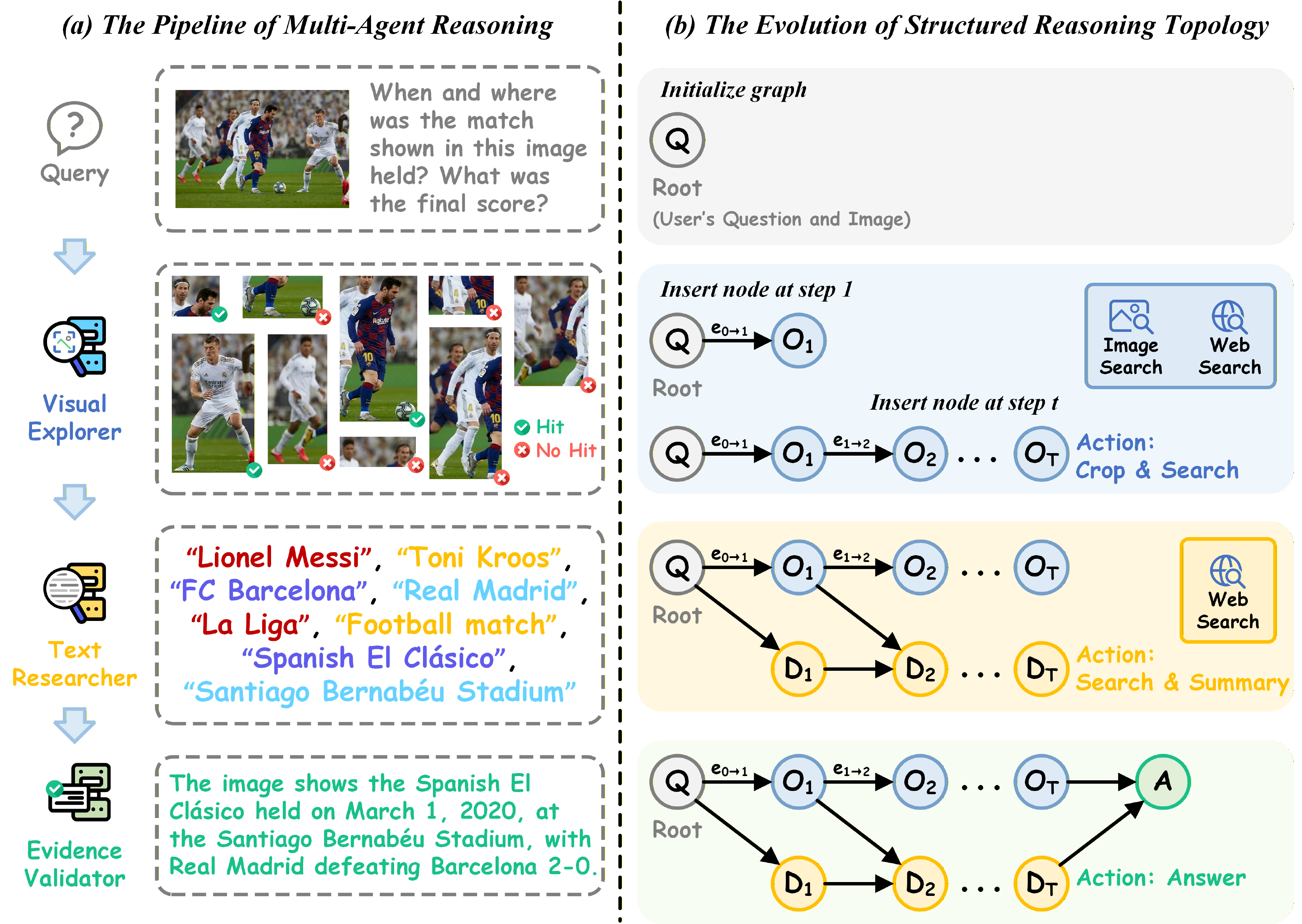}
    \caption{Overview of the GraMRAG framework.}
    \label{fig:gramrag_framework}
\end{figure*}

\section{Method}
\subsection{Vision-Text Bridged Reasoning Paradigm}

\paragraph{Multi-Scale Visual Entity Grounding.}
Given an input image $I$, a user query $q$, and a visual induction prompt $p_v$, the Visual Explorer performs hierarchical region proposal generation at each reasoning step $t$. Rather than submitting the full image as a single retrieval unit, which is prone to failure under background noise interference, the Visual Explorer simultaneously generates coarse-level contextual regions and fine-grained entity-level patches, forming a candidate bounding-box set $\mathcal{B}_t = \{b_t^1,\ b_t^2,\ \ldots,\ b_t^{N_t}\}$, where each $b_t^i$ represents a region crop at a distinct spatial scale or entity granularity, and $N_t$ denotes the number of proposals at step $t$. This candidate set induces a visual retrieval action $A_t^v = \mathrm{VSearch}(\mathcal{B}_t)$, which is processed sequentially by a two-stage visual toolchain: an image search module that matches each cropped region against a web-scale index and returns candidate URLs and a webpage retrieval module that fetches page content in structured markdown format.
The observation obtained at step $t$ is denoted:
\begin{equation}
    O_t^v = \mathrm{VisChain}(A_t^v)
    \label{eq:obs}
\end{equation}
Upon completing the current step, the Visual Explorer applies the induction prompt $p_v$ again to generate the next search context, producing the visual reasoning trajectory:
\begin{equation}
    \begin{split}
        \tau_\mathrm{vis} = \{&I,\ q,\ p_v,\ R_1,\ A_1^v,\ O_1^v,\\
        &\ldots,\ p_v,\ R_{T_v},\ A_{T_v}^v,\ O_{T_v}^v\}
    \end{split}
    \label{eq:vis_traj}
\end{equation}
where $R_t$ denotes the chain-of-thought reasoning generated prior to action $A_t^v$, and $T_v$ is the total number of visual retrieval steps.

Following each visual retrieval action, an independent Hit/No~Hit Judge evaluates the observation $O_t^v$ obtained at the current step, conditioned on the original image $I$ and query $q$, outputs a binary acceptance signal:
\begin{equation}
    s_t = \mathrm{Judge}(I,\ q,\ O_t^v) \in \{0, 1\}
    \label{eq:judge}
\end{equation}
When $s_t = 1$ (Hit), the retrieval is considered successful, meaning the retrieved content is semantically consistent with the query entity, and $O_t^v$ is added into the accumulated valid evidence pool:
\begin{equation}
    \mathcal{E}^v = \{O_t^v \mid s_t = 1,\ t = 1,\ \ldots,\ T_v\}
    \label{eq:ev_pool}
\end{equation}
When $s_t = 0$ (No Hit), the retrieval is considered unsuccessful, triggering the Visual Explorer to re-crop at a finer spatial granularity or to reformulate the search query before initiating a new retrieval attempt. 
The visual exploration phase terminates when $|\mathcal{E}^v|$ reaches a predefined sufficiency threshold or when the total step count reaches $T_v$. Decoupling the Hit/No~Hit judgment from the main agents and assigning it to an independent judge achieves a clean separation between exploration strategy and evidence evaluation, enhancing the robustness of the multi-scale iterative retrieval process.

\paragraph{Modality Bridging and Textual Reasoning.}
To transfer the evidence accumulated during the visual exploration phase into the textual reasoning domain, we introduce a modality bridging step. A captioning module first generates a comprehensive textual description $D$ of the input image $I$, covering key entities, spatial relationships, and scene-level semantics. The visual trajectory $\tau_\mathrm{vis}$ is then transformed into a bridged context: $I$ is replaced by $D$, all visual induction prompts $p_v$ are removed, while all chain-of-thought reasoning $R_t$, actions $A_t^v$, and judge-verified observations (i.e., entries in $\mathcal{E}^v$) are preserved. The bridged context is forwarded to the Text Researcher, a deep research LLM equipped with web search tools, which generates the subsequent textual reasoning trajectory:
\begin{align}
    \tau_\mathrm{text} = \{&D,\ q,\ R_1,\ A_1^v,\ O_1^v,\ \ldots,\ R_{T_v},\ A_{T_v}^v,\notag\\
    &O_{T_v}^v,\ R_{T_v+1},\ A_{T_v+1}^t,\ O_{T_v+1}^t,\notag\\
    &\ldots,\ R_{T_v+T_t},\ A_{T_v+T_t}^t\}
    \label{eq:text_traj}
\end{align}
where $T_t$ denotes the number of textual reasoning steps executed by the Text Researcher.

\paragraph{Evidence Aggregation and Complete Trajectory Construction.}
The Evidence Validator aggregates and verifies the complete collection of evidence produced across both reasoning phases.
Taking as input all judge-verified visual observations in $\mathcal{E}^v$ and the textual reasoning trajectory $\tau_\mathrm{text}$ produced by the Text Researcher, the Evidence Validator performs a cross-modal reasoning pass to resolve evidence conflicts, filter hallucinated content, and generate the final answer $a_\mathrm{out}$.
By grounding its reasoning in the graph-structured memory rather than a flattened context window, the Evidence Validator avoids the state-blindness inherent in linear memory paradigms, ensuring that the complete body of evidence across both modalities is faithfully incorporated into the final answer. The complete multimodal deep research trajectory is obtained by merging $\tau_\mathrm{vis}$ and $\tau_\mathrm{text}$:
\begin{align}
    \tau = \{&I,\ q,\ R_1,\ A_1^v,\ O_1^v,\ \ldots,\ R_{T_v},\ A_{T_v}^v,\notag\\
    &O_{T_v}^v,\ R_{T_v+1},\ A_{T_v+1}^t,\ O_{T_v+1}^t,\notag\\
    &\ldots,\ R_{T_v+T_t},\ A_{T_v+T_t}^t,\ a_\mathrm{out}\}
    \label{eq:full_traj}
\end{align}

\subsection{Multimodal Memory Graph}
\label{subsec:multimodal_memory_graph}

\begin{figure*}[t]
    \centering
    \includegraphics[width=\textwidth]{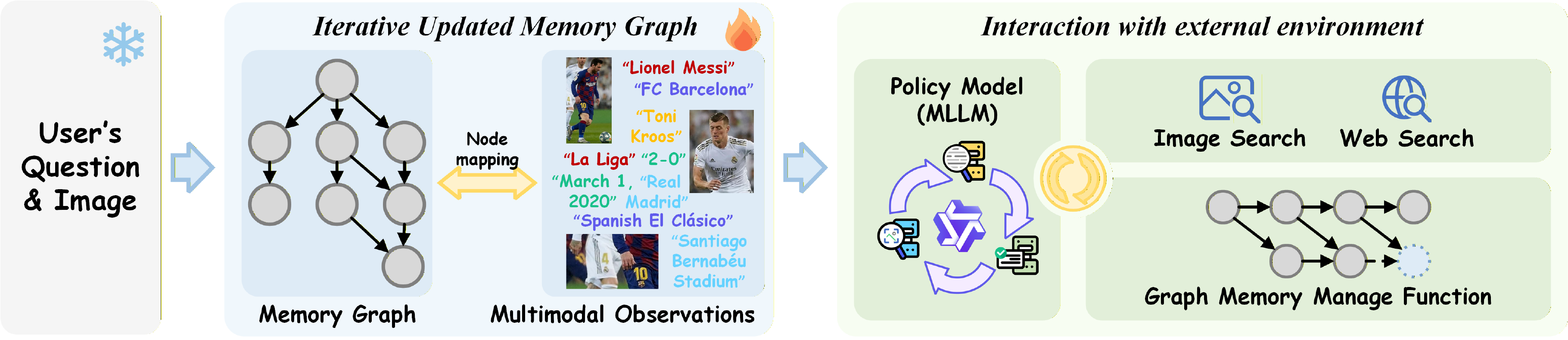}
    \caption{Inference pipeline of the GraMRAG framework.}
    \label{fig:inference_pipeline}
\end{figure*}

We construct a multimodal memory graph $\mathcal{G}_t = (\mathcal{V}_t,\, \mathcal{E}_t)$ that formalizes the multi-agent reasoning process as a dynamic DAG. This graph explicitly encodes action-observation dependencies across both the visual retrieval and textual research phases via a topological structure, fundamentally mitigating the state blindness inherent in linear memory paradigms and providing the structured foundation required by the TAPO algorithm. As illustrated in Figure~\ref{fig:gramrag_framework}(b), the memory graph evolves incrementally throughout reasoning, faithfully recording the exploration trajectory of each agent through node insertions and edge connections at each step.

\paragraph{Node Schema and Modality Classification.}
Each node $n_i \in \mathcal{V}_t$ encodes a discrete reasoning unit as a typed quadruple:
\begin{equation}
    n_i \triangleq \bigl(\mu_i,\ \kappa_i,\ \sigma_i,\ \nu_i\bigr)
    \label{eq:node}
\end{equation}
where $\mu_i \in \{n_{\text{root}},\, n_{\text{vis}},\, n_{\text{brd}},\, n_{\text{txt}},\, n_{\text{ans}}\}$ is the modality type tag; $\kappa_i$ is the search query or action descriptor driving the creation of this node; $\sigma_i$ is a concise textual summary of retrieved or generated content; and $\nu_i$ is the visual feature bank, populated only for $n_{\text{vis}}$-type nodes and set to empty for all other types. Five node types constitute the full reasoning pipeline. The root node ($n_{\text{root}}$) is initialized with user query $q$ and input image $I$, serving as the global anchor of the reasoning graph and permanently residing in $\mathcal{V}_t$. Visual nodes ($n_{\text{vis}}$) are spawned by the Visual Explorer at each retrieval step $t \in [1, T_v]$, storing the Hit/No~Hit judgment and the corresponding evidence summary, with $\nu_i$ preserving multimodal content from matched web pages. The bridge node ($n_{\text{brd}}$) is a singleton node representing the modality bridging step, storing image description $D$ as the semantic interface between the two reasoning phases. Text nodes ($n_{\text{txt}}$) are spawned by the Text Researcher at each retrieval step $t \in [T_v+1, T_v+T_t]$, storing web-retrieved textual evidence summaries. The answer node ($n_{\text{ans}}$) is spawned by the Evidence Validator, storing the final generated answer $a_\mathrm{out}$, whose insertion marks the termination of the reasoning process. The edge set $\mathcal{E}_t$ connects temporally adjacent nodes , encoding the sequential dependency of the reasoning flow. 

\paragraph{Iterative Graph Expansion Protocol.}
The memory graph $\mathcal{G}_t$ evolves through sequential node insertions jointly driven by the three main agents and the Hit/No~Hit Judge. The expansion protocol consists of four phases as follows. During visual expansion ($t \in [1, T_v]$), the Visual Explorer executes retrieval action $A_t^v = \mathrm{VSearch}(\mathcal{B}_t)$ and the Hit/No~Hit Judge outputs binary signal $s_t$; a visual node $n_t = (n_{\text{vis}},\, \kappa_t,\, \sigma_t,\, \nu_t)$ is then inserted, where $\sigma_t$ encodes the evidence summary alongside the $s_t$ judgment, $\nu_t$ preserves visual content from matched web pages, and a sequential edge $(n_{t-1}, n_t)$ connects it to its predecessor. During modality bridging, upon termination of the visual phase, the bridge node $n_{\text{brd}} = (n_{\text{brd}},\, \varnothing,\, D,\, \varnothing)$ is inserted and connected to the final visual node via the cross-modal bridge edge $(n_{T_v}, n_{\text{brd}})$, completing the semantic context transition from the visual to the textual domain. During textual expansion ($t \in [T_v+1, T_v+T_t]$), the Text Researcher executes web retrieval actions; text nodes $n_t = (n_{\text{txt}},\, \kappa_t,\, \sigma_t,\, \varnothing)$ are inserted with sequential edges, and the bridge node is connected to the first text node via edge $(n_{\text{brd}}, n_{T_v+1})$, fully closing the cross-modal transition path. During answer generation, once the Evidence Validator determines that the accumulated evidence is sufficient, it inserts the answer node $n_{\text{ans}} = (n_{\text{ans}},\, \varnothing,\, a_\mathrm{out},\, \varnothing)$ and the reasoning process terminates. 
Upon completion of all four phases, the memory graph captures the complete multimodal reasoning trajectory as a structured DAG rather than compressing it into a one-dimensional sequence, thereby enabling each agent to efficiently query the full historical reasoning state and simultaneously providing the topological backbone required for fine-grained credit assignment in the reinforcement learning phase.

\subsection{Topology-Aware Policy Optimization}

\begin{figure*}[t]
    \centering
    \includegraphics[width=\textwidth]{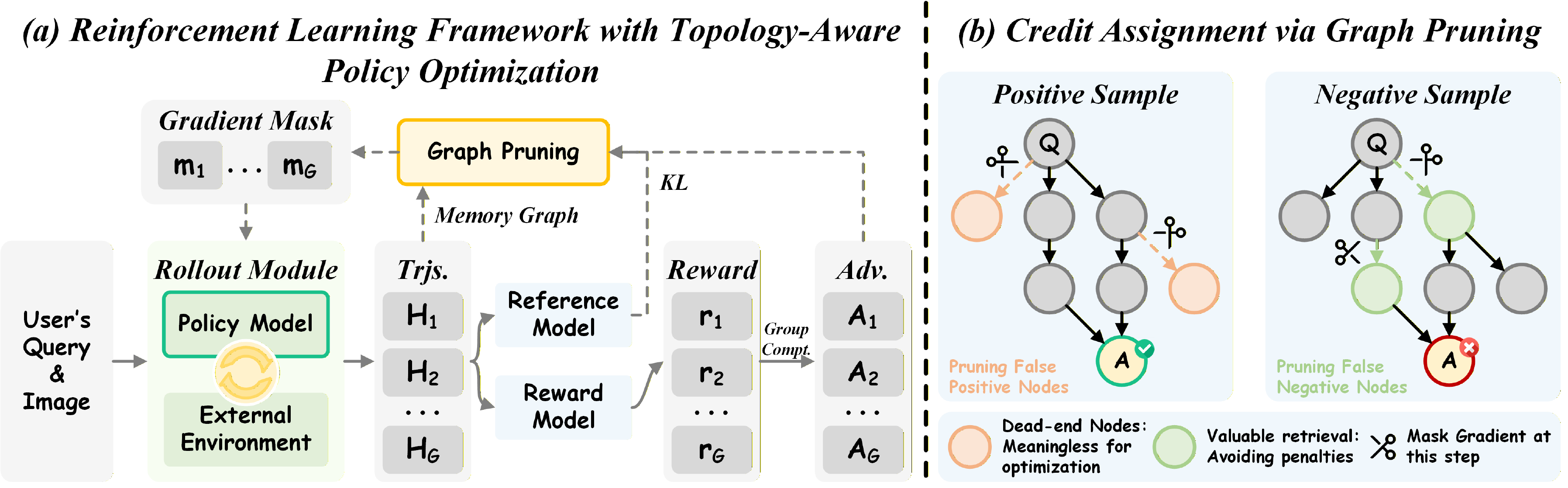}
    \caption{Overview of Topology-Aware Policy Optimization.}
    \label{fig:tapo}
\end{figure*}

As illustrated in Figure~\ref{fig:tapo}, we propose Topology-Aware Policy Optimization (TAPO) to address the systematic credit assignment problem in sparse reward reinforcement learning training, leveraging the topological structure of the memory graph to perform selective gradient masking that simultaneously rectifies both forms of bias across positive and negative trajectories with step-level precision.

\paragraph{Graph-Structured Trajectory Decomposition.}
We ground the reinforcement learning framework in the memory graph structure defined in Section~\ref{subsec:multimodal_memory_graph}. At each reasoning step $t$, the policy model $\pi_\theta$ receives the graph-serialized context:
\begin{equation}
    C_t = \{s_{\mathrm{sys}},\; q,\; H_t\}
    \label{eq:context}
\end{equation}
where $s_{\mathrm{sys}}$ is the system instruction and $H_t = \mathrm{Serialize}(\mathcal{G}_{t-1})$ is the graph-linearized memory context.
The policy model produces an reasoning segment that corresponds precisely to the construction of a single memory graph node $n_t$:
\begin{equation}
    s_t = (C_t,\; \zeta_t,\; A_t,\; O_t) \;\longrightarrow\; n_t
    \label{eq:segment}
\end{equation}
where $\zeta_t$ denotes the chain-of-thought reasoning preceding action $A_t$, and $O_t$ is the environmental observation returned by the external tools. The terminal segment takes the form $s_\mathrm{ans} = (C_\mathrm{ans},\; \zeta_\mathrm{ans},\; a_\mathrm{out})$. A complete rollout trajectory decomposes into an ordered sequence of segments $\Psi = \{s_1,\; s_2,\; \ldots,\; s_{T_v+T_t},\; s_\mathrm{ans}\}$. This node-based decomposition establishes a direct structural correspondence between policy gradient signals and graph topology: each gradient signal maps precisely to a specific graph node, providing the basis for the topology-driven credit assignment.

\paragraph{Topology-Driven Credit Assignment.}
For each rollout, we assign a binary outcome reward $r \in \{0, 1\}$ based on whether $a_\mathrm{out}$ matches the ground-truth answer $a_\mathrm{true}$. We then exploit the DAG topology of $\mathcal{G}_T$ to perform targeted gradient masking under two distinct scenarios as illustrated in Figure~\ref{fig:tapo}(b). Given a correct rollout ($r = 1$), we identify the critical reasoning path $\rho^*$ by traversing $\mathcal{G}_T$ backwards from $n_\mathrm{ans}$ to $n_\mathrm{root}$ along the directed edges. Nodes absent from $\rho^*$ constitute topological dead ends, namely exploratory steps that contribute no causal evidence to the correct answer. Propagating positive reward signals to these nodes would erroneously reinforce aimless exploration, so we suppress their gradient contributions. Given an incorrect rollout ($r = 0$), we identify the set of epistemically valuable nodes $\mathcal{N}_\mathrm{val} \subseteq \mathcal{V}_T$. For $n_{\text{vis}}$-type nodes, a node $n_t$ is designated as valuable when the Hit judgment $s_t$ stored within $\sigma_t$ records $s_t = 1$, confirming that the visual retrieval successfully located semantically relevant content. For $n_{\text{txt}}$-type nodes, a dedicated relevance evaluator further determines whether the textual evidence retrieved at step $t$ is directly relevant to the query. Excluding $\mathcal{N}_\mathrm{val}$ from the negative policy gradient update prevents the model from being penalized from performing demonstrably productive retrievals on trajectories that ultimately fail due to shortcomings in other reasoning steps.

Integrating both scenarios, we define the gradient mask $\psi_t \in \{0, 1\}$ for segment $s_t$ as:
\begin{equation}
    \psi_t = \mathbb{I}(r=1) \cdot \mathbb{I}(n_t \notin \rho^*) + \mathbb{I}(r=0) \cdot \mathbb{I}(n_t \in \mathcal{N}_\mathrm{val})
    \label{eq:mask}
\end{equation}
where $\mathbb{I}(\cdot)$ is the indicator function; $\psi_t = 1$ suppresses the gradient contribution of segment $s_t$ during the policy update, while $\psi_t = 0$ permits normal gradient update. The TAPO objective is defined as:
\begin{align}
    &\mathcal{J}_\mathrm{TAPO}(\theta) = \mathbb{E}\!\left[ \frac{1}{\sum_{g=1}^{G} n_g} \sum_{g=1}^{G}\sum_{i=1}^{n_g} (1 - \psi_{g,i})\right. \notag\\
    &\left.\cdot \mathcal{L}_\mathrm{clip}\!\left(\omega_{g,i}(\theta),\, \hat{A}_{g,i}\right) - \beta \cdot \mathrm{KL}\!\left[\pi_\theta \,\|\, \pi_\mathrm{ref}\right] \right]
    \label{eq:tapo}
\end{align}
where $\omega_{g,i}(\theta)$ is the probability ratio of segment $s_g^{(i)}$ under the current and old policies; $n_g$ denotes the number of segments in the $g$-th rollout. The $(1-\psi_{g,i})$ factor selectively gates gradient contributions based on the topological analysis of the memory graph: in positive trajectories, topological dead-end steps are masked to prevent their gradients from reinforcing redundant exploration; in negative trajectories, valuable retrieval steps are masked to prevent their gradients from penalizing effective retrieval behavior. Through this design, TAPO achieves precise step-level credit assignment across the entire multi-step reasoning trajectory, significantly outperforming conventional sparse-reward methods that rely solely on terminal outcome signals.

\section{Experiments}
\subsection{Datasets and Evaluation Metrics}

\begin{table*}[t]
    \small
    \centering
    \resizebox{1.00\textwidth}{!}{
    \begin{tabular}{l|cc|ccc|cccc|c}
    \toprule
    \multirow{2}{*}{\textbf{Method}} & 
    \multicolumn{2}{c|}{\textbf{General Text}} &
    \multicolumn{3}{c|}{\textbf{Image \& Visual Document}} &
    \multicolumn{4}{c|}{\textbf{Large-Scale Long-Context Video Corpus}} &
    \multicolumn{1}{c}{\multirow{2}{*}{\textbf{Overall}}} \\
    & \textbf{HotpotQA} & \textbf{SQuAD} & \textbf{ViDoSeek} & \textbf{SlideVQA} &
    \textbf{MMLongBench} & \textbf{LVBench} & \textbf{WikiHowQA} & \textbf{SyntheticQA} & \textbf{XVBench} & \\
    \midrule
    \multicolumn{11}{c}{$\textit{Qwen3-VL-4B-Instruct}$}\\
    \midrule
    ReAct & 62.8 & 63.9 & 40.7 & 45.8 & 13.4 & 13.0 & 13.2 & 29.1 & 22.5 & 33.8 \\
    Vanilla RAG & 62.4 & 63.2 & 44.9 & 43.4 & 14.9 & 12.8 & 12.1 & 33.7 & 27.1 & 34.9 \\
    UniversalRAG & 51.7 & 65.3 & 40.3 & 14.6 & 5.2 & 14.9 & 3.8 & 22.9 & 6.9 & 25.1 \\
    VideoRAG & 56.4 & 63.1 & 41.5 & 34.9 & 16.0 & 18.1 & 20.5 & 44.3 & 28.3 & 35.9 \\
    MemAgent & 66.2 & 70.3 & 46.4 & 42.7 & 10.9 & 20.1 & 20.2 & 33.4 & 23.7 & 37.1 \\
    Mem1 & 70.9 & 66.2 & 45.4 & 50.7 & 26.5 & 17.8 & 13.0 & 41.8 & 30.8 & 40.3 \\
    VimRAG & 75.7 & 74.0 & 48.3 & 52.9 & 28.3 & 23.5 & 21.5 & 52.4 & 33.5 & 45.6 \\
    \rowcolor{gray!15} \textbf{GraMRAG} & \textbf{78.4} & \textbf{77.1} & \textbf{51.2} & \textbf{55.6} & \textbf{30.8} & \textbf{26.0} & \textbf{24.1} & \textbf{57.3} & \textbf{36.0} & \textbf{48.5} \\
    \midrule
    \multicolumn{11}{c}{$\textit{Qwen3-VL-8B-Instruct}$}\\
    \midrule
    ReAct & 70.7 & 64.7 & 41.3 & 50.4 & 14.7 & 15.8 & 21.7 & 34.2 & 25.1 & 37.6 \\
    Vanilla RAG & 63.4 & 65.3 & 47.0 & 49.0 & 17.4 & 15.6 & 14.8 & 37.7 & 29.7 & 37.8 \\
    UniversalRAG & 56.3 & 66.7 & 45.2 & 16.9 & 6.9 & 20.4 & 10.7 & 24.0 & 8.7 & 28.4 \\
    VideoRAG & 62.5 & 62.3 & 43.1 & 36.0 & 19.3 & 23.8 & 27.1 & 48.6 & 32.1 & 39.4 \\
    MemAgent & 72.3 & 74.1 & 48.1 & 44.8 & 13.5 & 23.0 & 23.4 & 36.8 & 28.3 & 40.5 \\
    Mem1 & 71.5 & 68.6 & 44.2 & 57.0 & 34.0 & 22.5 & 20.8 & 44.0 & 31.4 & 43.8 \\
    VimRAG & 78.6 & 76.6 & 53.3 & 63.1 & 34.7 & 23.2 & 29.0 & 54.5 & 37.7 & 50.1 \\
    \rowcolor{gray!15} \textbf{GraMRAG} & \textbf{81.3} & \textbf{79.8} & \textbf{56.1} & \textbf{66.4} & \textbf{38.5} & \textbf{26.2} & \textbf{33.4} & \textbf{59.0} & \textbf{41.2} & \textbf{53.5} \\
    \bottomrule
    \end{tabular}
    }
    \caption{Performance comparison across different benchmarks.}
    \label{tab:overall_performance}
\end{table*}

We evaluate GraMRAG on a comprehensive set of benchmarks covering diverse task categories: general text benchmarks including HotpotQA~\cite{yang2018hotpotqa} and SQuAD~\cite{rajpurkar2016squad}; image and visual document benchmarks including ViDoSeek~\cite{wang2025vidorag}, SlideVQA~\cite{tanaka2023slidevqa} and MMLongBench~\cite{ma2024mmlongbench}; large-scale long-context video corpus benchmarks including LVBench~\cite{wang2025lvbench}, WikiHowQA and SyntheticQA~\cite{jeong2025videorag}, and XVBench~\cite{wang2026vimrag}. We employ a unified binary model-based metric ($\text{0}$ or $\text{1}$) across all tasks, where a reward model performs semantic matching between the generated and reference answers to ensure consistent and fair evaluation.

\subsection{Implementation Details}
We conduct experiments using Qwen3-VL-4B-Instruct and Qwen3-VL-8B-Instruct~\cite{bai2025qwen3} as the agent backbones. We use Qwen3-Max~\cite{yang2025qwen3} as the reward model to evaluate the quality and relevance of generated responses. Training is divided into two stages. In the SFT stage, we employ LlamaFactory~\cite{zheng2024llamafactory} with LoRA~\cite{hu2022lora} of rank 32 to fine-tune the language model while keeping the vision tower and multimodal projector frozen. Training spans 3 epochs with a batch size of 8 and gradient accumulation steps of 8, a learning rate of \(\text{1.0}\text{e}\text{-4}\), and a cosine learning rate scheduler. In the RL stage, we use rLLM~\cite{tan2025rllm} with the TAPO objective, an actor learning rate of \(\text{1.0}\text{e}\text{-6}\), a total of 2 training epochs, 8 agent groups, and a batch size of 32. All experiments are conducted on 8 NVIDIA H100 GPUs.

\subsection{Comparison with State-of-the-Art Methods}
We compare GraMRAG against representative baselines: ReAct~\cite{yao2022react} iteratively invokes external retrieval tools in a think-then-act paradigm; Vanilla RAG uses the original question as a query and performs direct MLLM inference over retrieved results; UniversalRAG~\cite{yeo2025universalrag} formulates cross-modal retrieval as a unified routing problem; VideoRAG~\cite{jeong2025videorag} extracts significant visual information via key-frame selection; MemAgent~\cite{yu2025memagent} manages context by sequentially feeding search results to the model; Mem1~\cite{zhou2025mem1} dynamically compresses agent memory through a cyclical retrieval-then-memorization process; and VimRAG~\cite{wang2026vimrag} models reasoning process to manage large-scale multimodal contexts.

As presented in Table~\ref{tab:overall_performance}, linear history accumulation paradigms such as ReAct rapidly exhaust the context window under the heavy token overhead of visual data; task-specific methods including VideoRAG and UniversalRAG exhibit limited generalizability due to fixed inference pipelines; and summarization-based memory paradigms such as MemAgent and Mem1, lacking explicit modeling of agent reasoning states, progressively lose track of historical retrieval paths as interaction rounds accumulate; and VimRAG models the reasoning process but lacks long-horizon reasoning capability. Our GraMRAG achieves consistently state-of-the-art performance across all benchmarks, achieving overall average scores of 48.5\% and 53.5\% under the Qwen3-VL-4B-Instruct and Qwen3-VL-8B-Instruct settings respectively, validating the effectiveness of formalizing the multi-step reasoning process as a dynamic DAG with topology-driven fine-grained credit assignment for multimodal knowledge-intensive reasoning tasks.

\subsection{Ablation Study}

\begin{table}[!t]
    \small
    \centering
    \resizebox{1.00\linewidth}{!}{
    \begin{tabular}{cc|cc|cc|c}
    \toprule
    \multicolumn{2}{c|}{\textbf{Reasoning Horizon}} & \multicolumn{2}{c|}{\textbf{Memory Paradigm}} & \multicolumn{2}{c|}{\textbf{Policy Optimization}} & \multirow{2}{*}{\textbf{Acc}} \\
    \textbf{Standard} & \textbf{Long} & \textbf{Iterative} & \textbf{Graph} & \textbf{GRPO}  & \textbf{TAPO} & \\
    \midrule
     \ding{51} & & \ding{51} & & & & 43.8 \\
     & \ding{51} & \ding{51} & & & & 44.2 \\
     & \ding{51} & & \ding{51} & & & 47.5 \\
     & \ding{51} & & \ding{51} & \ding{51} & & 49.3 \\
     \rowcolor{gray!15} & \ding{51} & & \ding{51} & & \ding{51} & \textbf{53.5} \\
    \bottomrule
    \end{tabular}
    }
    \caption{Ablation study across all benchmarks.}
    \label{tab:ablation}
\end{table}

To validate the contribution of each component in GraMRAG, we conduct ablation studies across all benchmarks.
As shown in Table~\ref{tab:ablation}, the three components exhibit clear progressive complementarity: long-horizon reasoning enables sustained multi-round retrieval across extended cross-modal reasoning trajectories; the multimodal memory graph demonstrates that formalizing reasoning as a dynamic DAG effectively mitigates state blindness and suppresses redundant retrieval; and replacing the standard sparse-reward method Group Relative Policy Optimization (GRPO)~\cite{shao2024deepseekmath} with TAPO delivers a further improvement, directly confirming the critical role of topology-guided credit assignment in fine-grained policy optimization. The complete model achieves an average score of 53.5\%, an improvement of over 9 points above the baseline, comprehensively validating the necessity and complementarity of each component.

\section{Analysis and Discussion}

\begin{figure}
    \centering
    \includegraphics[width=0.46\textwidth]{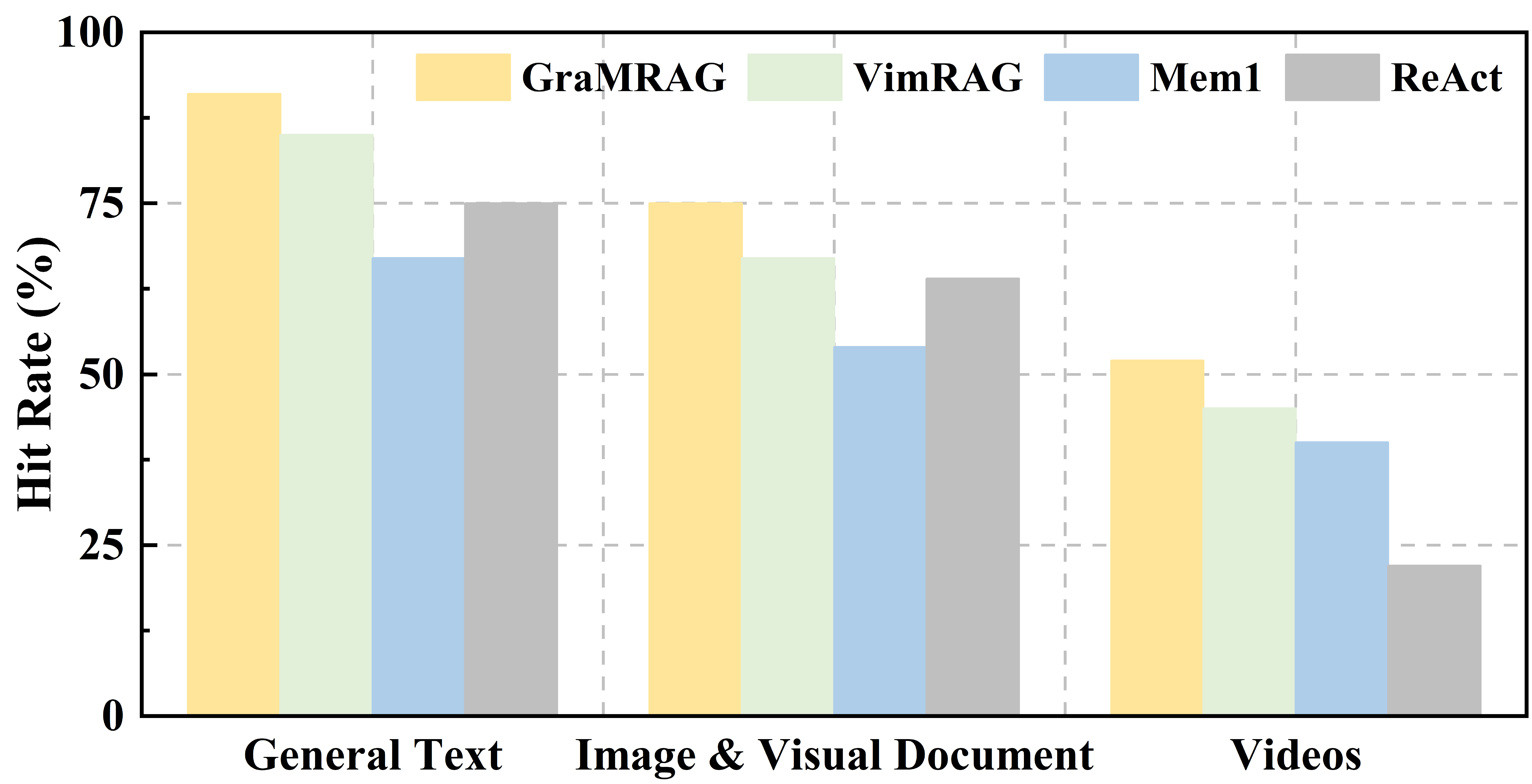}
    \caption{Analysis of retrieval performance.}
    \label{fig:retrieval_performance}
\end{figure}

\textbf{Retrieval quality depends on reasoning depth.} High-quality answer generation in complex multimodal tasks depends on sustaining precise multi-round retrieval across a long reasoning horizon. 
Traditional memory paradigms cause agents to progressively lose track of historical retrieval paths and accumulate redundant searches that severely constrain effective reasoning depth.
GraMRAG addresses this by transferring the iterative planning capability into the multimodal domain via the vision-text bridged reasoning paradigm, supporting trajectories spanning tens of steps; the multimodal memory graph further encodes action-observation dependencies explicitly, directing each retrieval attempt precisely toward unexplored evidence space. As shown in Figure~\ref{fig:retrieval_performance}, this design yields substantially higher hit rate across all three task categories, directly validating the decisive role of long-horizon reasoning depth in determining retrieval quality.

\textbf{Graph memory mitigates state blindness.} The dynamic DAG representation of the multimodal memory graph explicitly encodes action-observation dependencies as directed edges, enabling each agent to efficiently access the complete historical reasoning trajectory and effectively mitigating state blindness. As shown in Table~\ref{tab:ablation}, replacing iterative summarization-based memory with the multimodal memory graph yields a 3.3-point performance improvement, validating that the multimodal memory graph effectively mitigates state blindness and suppresses redundant retrieval, confirming the necessity of structured memory in long-horizon multimodal reasoning.

\begin{figure}
    \centering
    \includegraphics[width=0.46\textwidth]{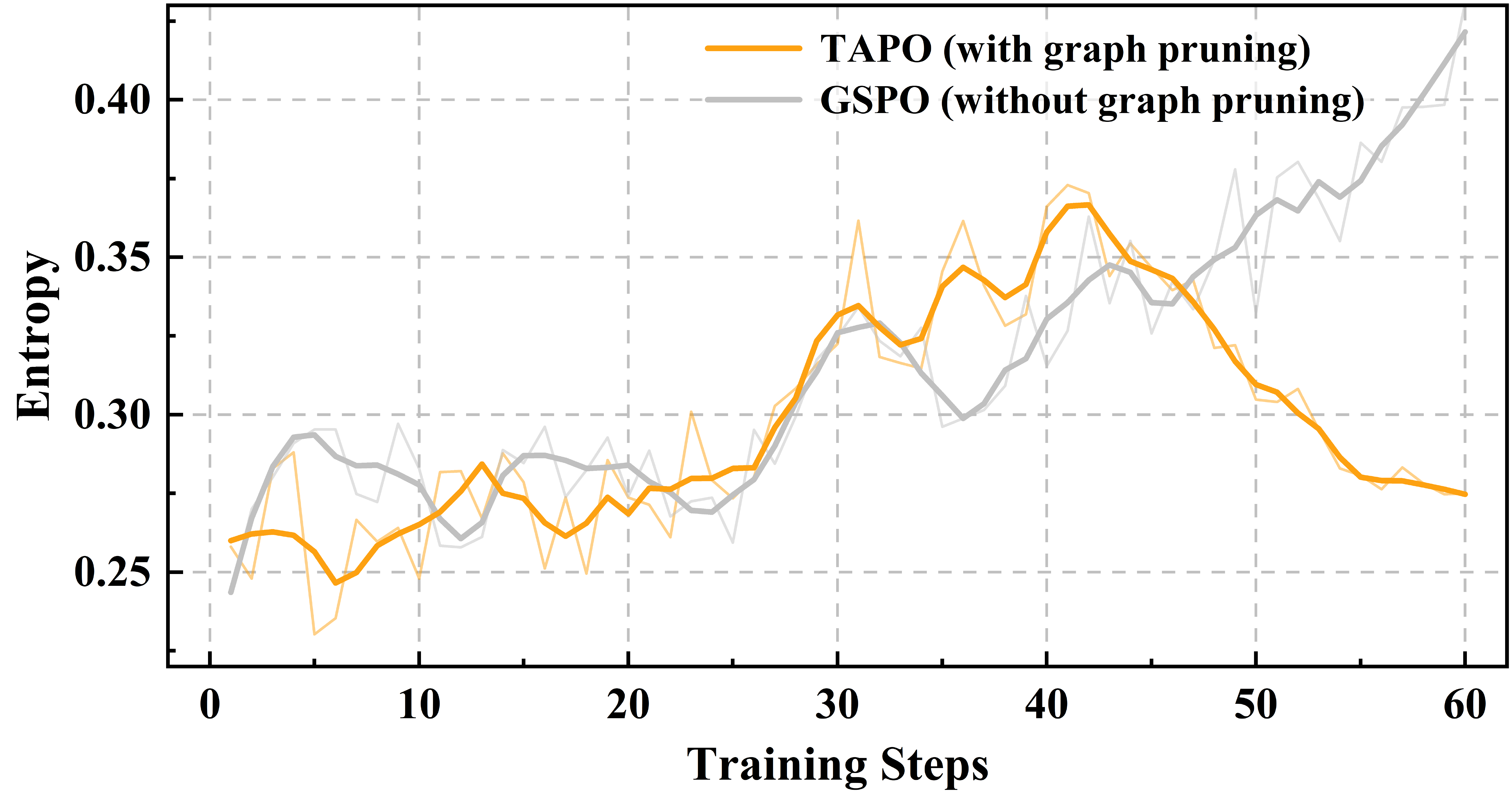}
    \caption{Analysis of training efficiency.}
    \label{fig:training_efficiency}
\end{figure}

\textbf{Graph pruning accelerates training convergence.} As demonstrated in Figure~\ref{fig:training_efficiency}, TAPO achieves faster convergence throughout training compared to the baseline Group Sequence Policy Optimization (GSPO)~\cite{zheng2025group}, indicating that topology-guided graph pruning drives the model toward faster convergence by supplying cleaner credit assignment signals. This yields two critical insights into agentic reinforcement learning: first, optimization stability fundamentally depends on ensuring correct positive gradient signals while eliminating ambiguous updates from negative trajectories; second, the quality of rollout samples, specifically their preference alignment, is more decisive for performance than simply scaling up the training set. 
By simultaneously masking dead-end nodes in positive trajectories and preserving valuable nodes in negative ones, TAPO achieves fine-grained step-level credit assignment rather than coarse terminal-reward-level optimization.

\section{Conclusion}

In this paper, we introduce GraMRAG, a novel graph memory-guided multi-agent RAG framework for complex multimodal long-horizon reasoning tasks. Our approach integrates three key innovations: a vision-text bridged reasoning paradigm that unifies multi-scale entity cropping with a ReAct-style visual toolchain for long-horizon cross-modal reasoning; multimodal memory graph construction that formalizes agent reasoning as a dynamic directed acyclic graph (DAG) to fundamentally mitigate state blindness and suppress redundant retrieval; and Topology-Aware Policy Optimization (TAPO) algorithm that leverages graph topology for fine-grained credit assignment across multi-step reasoning trajectories. Extensive experiments on diverse multimodal benchmarks demonstrate that our method consistently outperforms existing baselines and achieves state-of-the-art performance. Future work will investigate the scalability of the proposed framework to larger architectures and broader multimodal scenarios.

\section{Limitations}
Although GraMRAG achieves significant performance improvements on complex multimodal multi-step reasoning tasks, the multi-agent collaboration framework introduces some inference overhead compared to traditional RAG approaches. Notably, such a trade-off is generally acceptable in deep research scenarios where users prioritize reasoning accuracy over response speed. Nevertheless, it may hinder direct deployment in resource-constrained environments. Therefore, optimizing the interaction strategy among agents and designing more efficient lightweight graph memory mechanisms constitute key directions for future work.

\end{document}